\documentclass[hidelinks]{isprs}
\usepackage{subfigure}
\usepackage{setspace}
\usepackage{geometry} 
\usepackage{epstopdf}
\usepackage[labelsep=period]{caption}  

\usepackage{hyperref}
\usepackage{graphicx}
\usepackage[export]{adjustbox}
\usepackage{amsmath,amsfonts,amssymb}
\usepackage[acronyms]{glossaries}
\usepackage[sort]{natbib} 
	\renewcommand{\refname}{REFERENCES} 
\usepackage{color, soul}
\usepackage{diagbox}	
\usepackage{enumitem}
\usepackage{multirow}
\usepackage[nameinlink, capitalize]{cleveref}

\usepackage{xspace}

\newcommand{\wrt}{w.r.t.\ }
\newcommand{\m}{\,m\xspace}

\begin{document}

\title{Deep cross-domain building extraction for selective depth estimation from oblique aerial imagery}

\author{
B. Ruf\textsuperscript{a,b}, L. Thiel\textsuperscript{a}, M. Weinmann\textsuperscript{b}
}

\address{
\textsuperscript{a}Fraunhofer IOSB, Video Exploitation Systems, 76131 Karlsruhe, Germany -\\ \{boitumelo.ruf, laurenz.thiel\}@iosb.fraunhofer.de\\
\textsuperscript{b}Institute of Photogrammetry and Remote Sensing, Karlsruhe Institute of Technology, \\ 76131 Karlsruhe, Germany - \{boitumelo.ruf, martin.weinmann\}@kit.edu
}


\commission{I, }{I} 
\workinggroup{} 
\icwg{I/II}   

\keywords{aerial oblique imagery, object detection, building extraction, deep learning, convolutional neural networks, transfer learning, depth estimation, semi-global matching}

\newacronym{SfM}{SfM}{Structure-from-Motion}
\newacronym{WTA}{WTA}{Winner-Takes-It-All}
\newacronym{DEMs}{DEMs}{Digital Elevation Models}
\newacronym{SGM}{SGM}{semi-global matching}
\newacronym{GPGPU}{GPGPU}{general purpose computation on a GPU}
\newacronym{NTSB}{NTSB}{New Tsukuba Stereo Benchmark}
\newacronym{MVS}{MVS}{multi-view stereo}
\newacronym{RPN}{RPN}{region proposal network}
\newacronym{COTS}{COTS}{commercial off-the-shelf}
\newacronym{AP}{AP}{average precision}
\newacronym{mAP}{mAP}{mean average precision}
\newacronym{CNN}{CNN}{convolutional neural network}
\newacronym{CNNs}{CNNs}{convolutional neural networks}
\newacronym{ROI}{RoI}{region of interest}
\newacronym{ROIs}{RoIs}{regions of interest}
\newacronym{LSD}{LSD}{line segment detector}
\newacronym{NMS}{NMS}{non-maximum suppression}
\newacronym{SVM}{SVM}{support vector machine}
\newacronym{LOD}{LOD}{level of detail}

\abstract{
With the technological advancements of aerial imagery and accurate 3d reconstruction of urban environments, more and more attention has been paid to the automated analyses of urban areas. In our work, we examine two important aspects that allow online analysis of building structures in city models given oblique aerial image sequences, namely automatic building extraction with \gls*{CNNs} and selective real-time depth estimation from aerial imagery. We use transfer learning to train the Faster R-CNN method for real-time deep object detection, by combining a large ground-based dataset for urban scene understanding with a smaller number of images from an aerial dataset. We achieve an \gls*{AP} of about 80\% for the task of building extraction on a selected evaluation dataset. Our evaluation focuses on both dataset-specific learning and transfer learning. Furthermore, we present an algorithm that allows for multi-view depth estimation from aerial image sequences in real-time. We adopt the \gls*{SGM} optimization strategy to preserve sharp edges at object boundaries. In combination with the Faster R-CNN, it allows a selective reconstruction of buildings, identified with \gls*{ROIs}, from oblique aerial imagery.
}

\maketitle

\glsresetall 
\section{INTRODUCTION}
\label{sec:intro}

\sloppy

In recent years, more and more attention has been paid to the automated analyses of urban areas due to an increase in urbanization and the need for more efficient urban planing and sustainable development.
While the consideration of urban trees in planning processes can provide measurable economic, environmental, social and health benefits \citep{Kelly_2011}, the monitoring and analysis of the given buildings e.g. allows to create and update city models \citep{Kolbe_2009}, finding suitable roof planes for solar energy installations \citep{Schuffert_et_al_2015} and reasoning about a diversity of processes.

The basis for such automated analyses is typically given with data acquired from aerial platforms. The automated extraction of objects from such data has been a topic of great interest over decades \citep{Mayer_2008}. 
To reason about possible updates or extract objects of interest, the 2d outlines of buildings as well as a reconstruction of the 3d shape of buildings may be compared to the information given in model data. To foster research on both of these issues, the {ISPRS} benchmark on urban object classification and 3d building reconstruction \citep{Rottensteiner_et_al_2012} has been initialized and addressed by a diversity of approaches.

Due to technological advancements, aerial oblique imagery nowadays provides a ubiquitous tool for an accurate 3d reconstruction of urban environments \citep{Cavegn_et_al_2014}. Based on the 3d reconstruction, city models with different \gls*{LOD} can be created. Particularly \gls*{LOD}2 building models with distinctive roof structures and larger building installations like balconies and stairs and \gls*{LOD}3 building models with detailed wall and roof structures, doors and windows \citep{Kolbe_2009} are desirable for a variety of applications. Recent work in the field of 3d change detection \citep{Taneja2015Change,Ruf2016,Palazzolo2017change} suggests to use images for an efficient verification or improvement of such models. 

In our work, we aim at methodologies that allow online change detection and analysis of building structures in city models given oblique aerial image sequences captured from \gls*{COTS} drones. Hereby, we use ``online'' to indicate that our algorithms should allow a direct annotation and processing of the input sequence. A key aspect in this process is the real-time image-based depth estimation, giving us information on the current structure of the scene. However, the model data estimated from imagery typically differs greatly from city models in its detail and the objects it holds. While city models only depict buildings, image-based models also contain vegetation and dynamic objects such as cars and pedestrians. Apart from being potential sources for errors, the reconstruction of such objects is not necessary for finding changes in building structures.

In this paper, we examine two aspects to remedy this, namely extraction of buildings from aerial imagery and selective real-time image-based depth estimation for the identified objects. As no dataset exists that allows to train and test a method for building extraction from imagery captured by \gls*{COTS} drones, we use transfer learning by using a large ground-based dataset intended for semantic urban scene understanding and adopting a small aerial dataset to our needs. The assumption is that oblique imagery captured from low altitudes show building parts that are also depicted in ground-based imagery, which e.g. holds for fa\c{c}ades.\ \\

In summary, our main contribution is a methodology for building extraction and selective 3d reconstruction from oblique aerial imagery which
\begin{itemize}[topsep=0pt, partopsep=0pt, itemsep=0pt]
	\item relies on cross-domain training of a \gls*{CNN} to deliver object proposals corresponding to buildings,
    
    \item allows real-time image-based depth estimation that preserves strong discontinuities at object boundaries, giving buildings sharp edges, and
    \item is evaluated on different datasets, whereby the focus is put on both dataset-specific learning and transfer learning to obtain appropriate object proposals as the basis for 3d reconstruction.
\end{itemize}

This paper is structured as follows: We briefly summarize related work in Section~\ref{sec:related_works} giving an overview of the recent advances in the fields of object detection and image-based 3d reconstruction. In Section~\ref{sec:methodology}, we give a detailed description of our methodology for deep building extraction and depth estimation. The training process together with the datasets used, as well as the experimental results are described in Section~\ref{sec:eval}. We give a short discussion of the achieved results in Section~\ref{sec:discussion} before we conclude our paper in Section~\ref{sec:conclusion}.

\section{RELATED WORK}
\label{sec:related_works}

\sloppy

In the following, we provide an overview on the advances in the field of object detection, followed by a short outline on the related work of image-based 3d reconstruction.

\subsection{Object Detection}
\label{sec:RW_Instance_Search}

In the past, a rich variety of approaches have been presented for object detection in given imagery. Thereby, several approaches focus on instance search, i.e. the detection of specific types of objects by generating candidate windows around \gls*{ROIs} and thus delivering object proposals. In general, this can be achieved with \emph{window scoring methods} and \emph{grouping methods}. For a more detailed discussion of such methods, we refer to \citep{Sommer_et_al_2016}, and we instead only briefly summarize the main ideas.

The \emph{window scoring methods} typically generate candidate windows by applying either a sliding window approach or a random sampling technique. 
For each candidate window, a score is calculated which allows to rank or discard these windows \wrt their score. 
To define the scoring of candidate windows, different cues may be taken into account such as a generic objectness measure quantifying how likely it is for an image window to contain an object of any class \citep{Alexe_et_al_2012}. 
This measure combines several image cues such as multiscale saliency, color contrast, edge density or superpixels straddling. 
A different objectness measure relies on the number of edges that exist in the window and those edges that are members of contours overlapping the window's boundary \citep{Zitnick_and_Dollar_2014}.

The \emph{grouping methods} typically perform an image segmentation followed by a grouping of segments to generate multiple (possibly overlapping) segments that are likely to correspond to objects. Thereby, the grouping is typically based on a diverse set of cues including superpixel shape, appearance cues and boundary estimates \citep{Hosang_et_al_2016,Sommer_et_al_2016}. A commonly used approach is known as selective search \citep{Uijlings2013selective}. This approach relies on a hierarchical grouping to get a set of small starting regions forming the basis of a selective search. Subsequently, a greedy data-driven algorithm is used to iteratively merge regions based on a variety of complementary grouping criteria and a variety of complementary color spaces with different invariance properties. This yields a small set of high-quality object locations.

\glsunset{CNNs}
With the great success of modern deep learning in a variety of research domains, such techniques have also been introduced in recent years for object detection in the form of deriving object proposals. The combination of a region proposal method with deep \gls*{CNNs} has been proposed with the R-CNN \citep{Girshick2014RCNN}. This approach makes use of selective search \citep{Uijlings2013selective} to generate category-independent region proposals, but is generally agnostic to the particular region proposal method. The derived proposals are provided as input to a \gls*{CNN} that extracts a feature vector of fixed length for each region. Finally, the feature vectors characterizing the derived \gls*{ROIs} are used to classify the objects located within these regions using a \gls*{SVM}. The R-CNN approach outperformed classical methods on public benchmarks for object detection. To improve efficiency, the Fast R-CNN \citep{Girshick2015FastRCNN} has been proposed. Amongst others, the introduced innovations also allow using the very deep VGG-16 network \citep{Simonyan2014VGG16} which performed quite well in the ImageNet Localization+Classification Challenge 2014 \citep{Russakovsky2015ILSVRC}. At this point, the computation of region proposals was identified as bottleneck in terms of runtime.

Instead of using conventional approaches to derive region proposals \citep{Uijlings2013selective,Felzenszwalb2010object}, the use of a \gls*{RPN} has been proposed \citep{Ren2017FasterRCNN} which takes an image (of any size) as input and delivers a set of candidate windows, i.e. rectangular object proposals, each with an objectness score. These region proposals in turn are used by a Fast R-CNN for object detection and, due to the increase in computational efficiency, the resulting approach was dubbed Faster R-CNN \citep{Ren2017FasterRCNN}. This improvement allows real-time object detections on a GPU achieving state-of-the-art results of over $70\%$ \gls*{mAP} on public benchmarks. 

\begin{figure*}[!ht]%
	\centering
	\includegraphics[width=\textwidth]{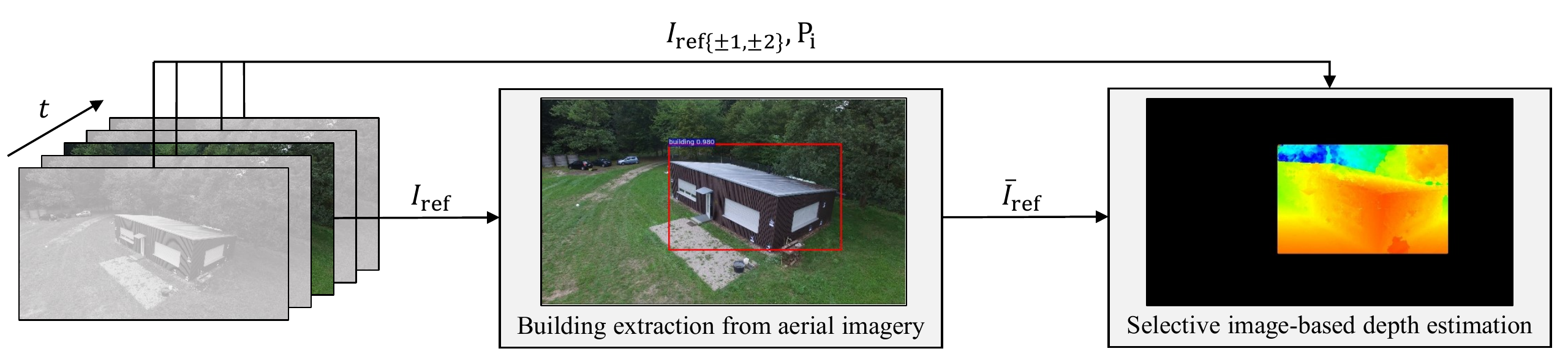}
	\caption{Overview of our methodology for building-aware selective image-based depth estimation.}
	\label{fig:overview}
\end{figure*}

\subsection{Image-based 3d Reconstruction}
\label{sec:RW_3d_recon}

In recent years, a lot of attention has been paid to dense image-based depth estimation and model reconstruction. With the ever increasing computational power of modern hardware, it is possible to perform an online reconstruction from imagery acquired with a single moving camera \citep{Newcombe2010,Newcombe2011,Stuehmer2012}. The presented 3d models are of great detail, but typically depict only a small-scale scenery. 

With \gls*{SGM} \citep{Hirschmueller2008}, an optimization strategy was introduced which can be used for depth estimation from a stereo setup or a multi-image matching \citep{Rothermel2012}. Since its announcement, \gls*{SGM} has been used and adopted for a wide variety of applications in the field of dense depth estimation and 3d reconstruction as it provides a good trade-off between accuracy and computational effort. As a result, state-of-the-art methods for aerial image-based 3d reconstruction also rely on the \gls*{SGM} optimization \citep{Rothermel2012,Angelo2012,Haala2015}.

\section{METHODOLOGY}
\label{sec:methodology}

\sloppy

The processing pipeline of our approach is composed of two parts: building extraction and selective depth estimation, as outlined in Figure \ref{fig:overview}. For each iteration of the presented pipeline, we choose a bundle of five consecutive images of an input sequence which depict the scene of interest from five slightly different viewpoints. We select the center image of the input bundle as our reference image and pass it on to the first processing step in which we use an object detection algorithm to identify buildings and mark these with axis-aligned \gls*{ROIs}. The annotated reference image is then passed on to the second step in which it is combined with the remaining four images of the input bundle to estimate the depth for the previously identified \gls*{ROIs}. In the following, both parts of the pipeline will be discussed in more detail.

\subsection{Building Extraction from Aerial Imagery}
\label{sec:meth_detection}

\begin{figure}[!b]%
	\centering
	\includegraphics[width=\columnwidth]{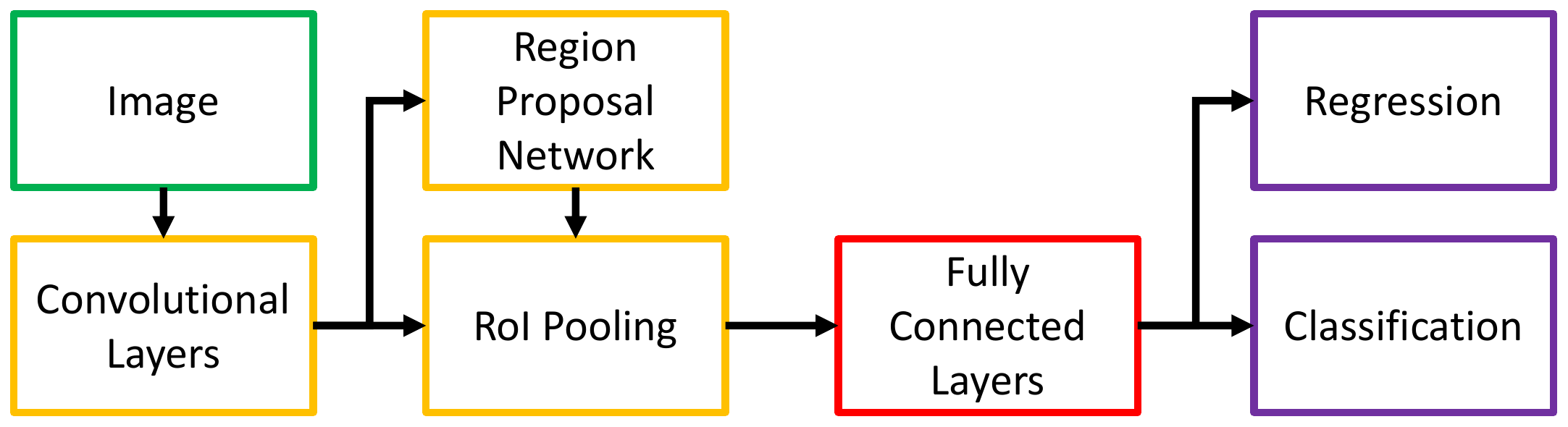}
	\caption{Structural overview of the Faster R-CNN. Based on the deep VGG-16 network, it is composed of 13 convolutional layers and two fully connected layers. The object proposals are classified by a softmax classifier and their coordinates are computed by the regression layer.}
	\label{fig:faster_rccn}
\end{figure}

\glsunset{ROI}
For the task of building extraction from aerial imagery, we have adopted and trained the Faster R-CNN \citep{Ren2017FasterRCNN} based on the VGG-16 network of \citet{Simonyan2014VGG16}. As depicted in Figure \ref{fig:faster_rccn}, the Faster R-CNN first passes the input image to a set of convolutional layers creating a feature map. The VGG-16 based Faster R-CNN holds 13 convolutional layers and produces a feature map of size $14\times 14\times 512$. The second processing step is made up of the \gls*{RPN} which computes numerous proposals by sliding differently sized windows over the input. As the \gls*{RPN} is sharing its convolutional layers with the rest of the network, it takes the previously computed feature map as input. The resulting proposals are recombined with the initial feature map in the third step and fed into a \gls*{ROI} pooling layer which uses max-pooling to convert the feature map within each proposal into a spatially confined feature map of size $7\times 7\times 512$. Two fully connected layers then produce feature vectors of size $1\times 1\times 4096$ which are passed to the bounding box regression to compute four coordinates for each of the bounding boxes and to the softmax classifier to compute the corresponding class scores. We have adjusted the output size of the classifier to two, as we only want to distinguish if an identified object is a building or not. As the regression layer computes the four coordinates of each bounding box, its output size is set to eight accordingly.

Furthermore, as we aim to detect buildings in oblique aerial imagery, which might result in a partial occlusion of buildings by other buildings, we have substituted the conventional \gls*{NMS} by the Soft-\gls*{NMS} as presented by \citet{Bodla2017SoftNMS}. Instead of removing all bounding boxes of an object that lie up to a certain threshold within the bounding box with the highest score $\mathcal{M}$, the Soft-\gls*{NMS} decays the score of all non-maximum boxes according to the amount of overlap \wrt $\mathcal{M}$. This results in better detections of objects which are partially occluded by other objects of their kind, as the proposals are not eliminated immediately.

\begin{figure*}[!ht]
  \centering
  \subfigure[]{\includegraphics[width=0.5\columnwidth,frame]{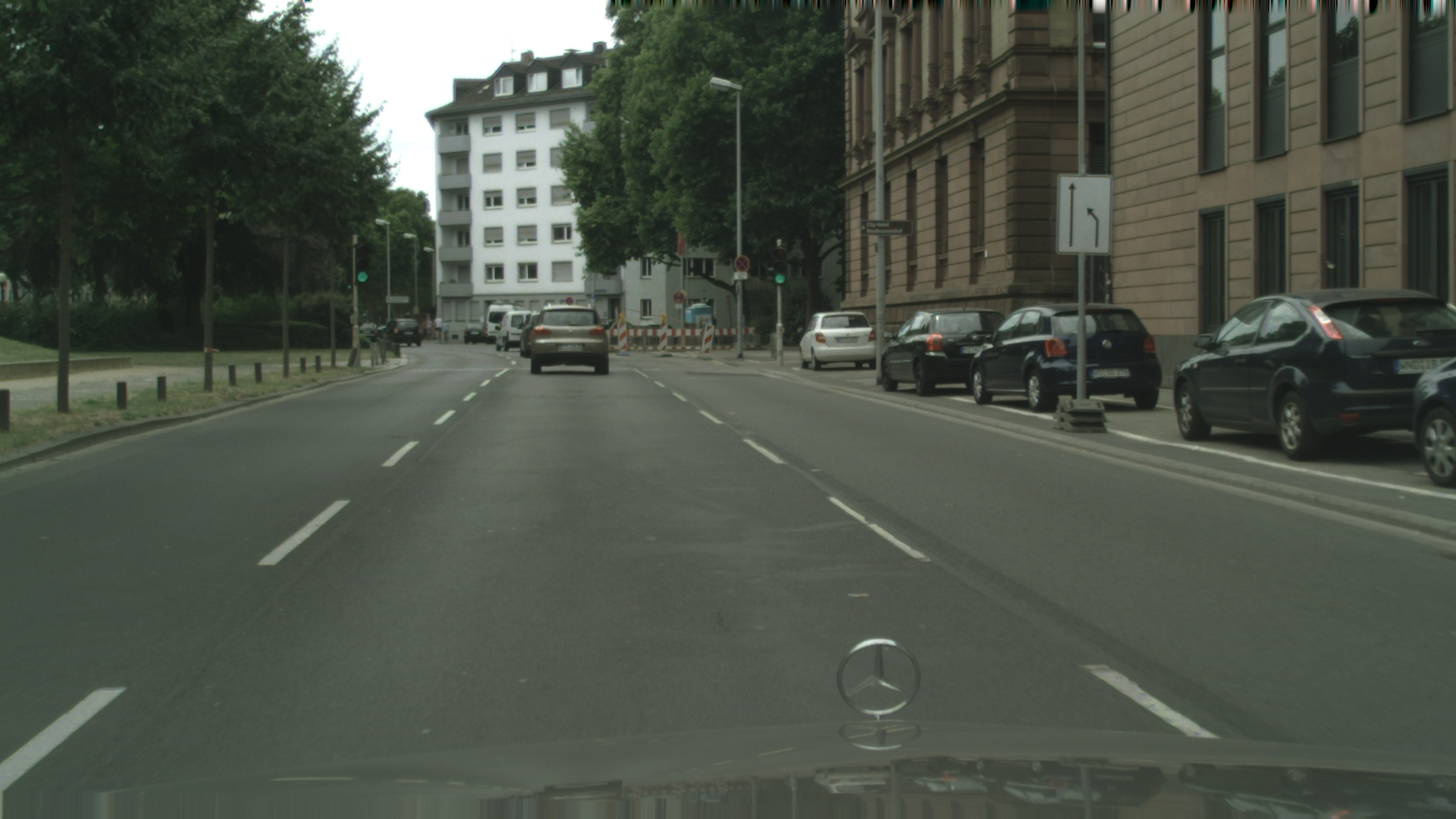}\label{fig:datasets_cityscape}} \hfill
  \subfigure[]{\includegraphics[width=0.5\columnwidth,frame]{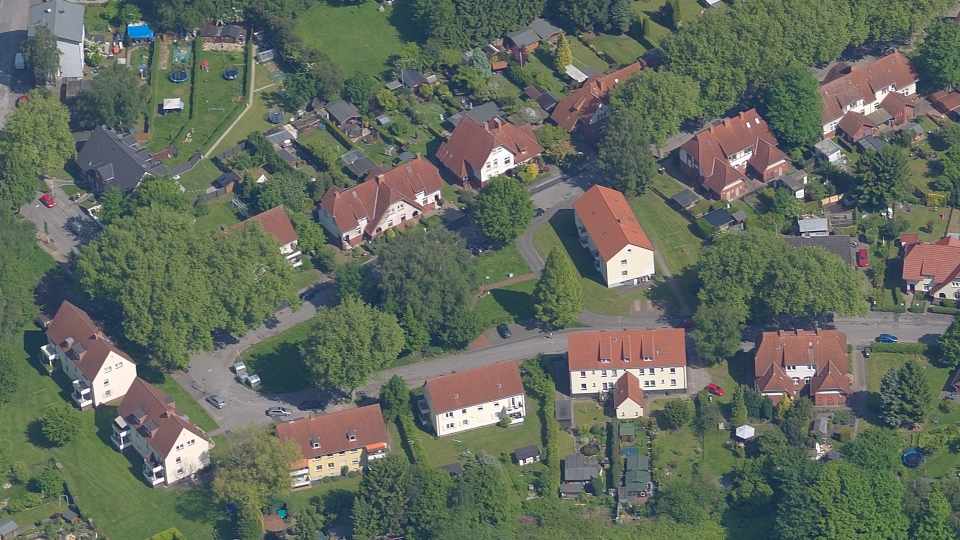}\label{fig:datasets_isprs}} \hfill
  \subfigure[]{\includegraphics[width=0.5\columnwidth,frame]{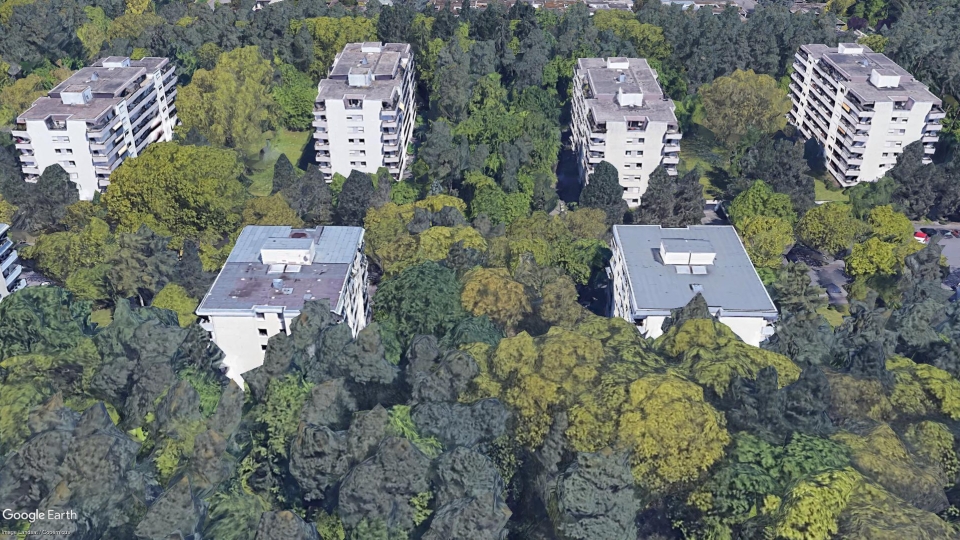}\label{fig:datasets_google}} \hfill
  \subfigure[]{\includegraphics[width=0.5\columnwidth,frame]{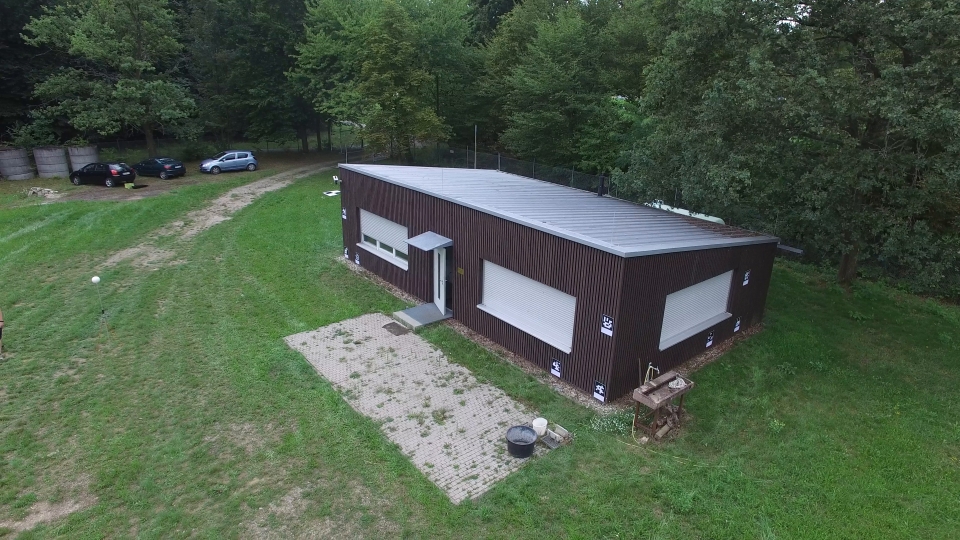}\label{fig:datasets_tmb}}
  \caption{Example figures of the datasets used. \textbf{(a)} Cityscapes dataset showing an urban street scene with building fa\c{c}ades. \textbf{(b)} ISPRS benchmark dataset comprising oblique aerial imagery of a suburban area. \textbf{(c)} Exemplary scene of the GoogleEarth dataset depicting different building kinds and dense vegetation. \textbf{(d)} Excerpt of our own dataset used to demonstrate the local image-based depth estimation.}
  \label{fig:datasets}
\end{figure*}

\subsection{Selective Image-based Depth Estimation}
\label{sec:meth_depth}

For image-based depth estimation, we employ a multi-view plane-sweep algorithm. As input, our algorithm takes a bundle of five consecutive images, one reference image $I_\mathrm{ref}$ and four matching images $I_\mathrm{ref\{\pm 1, \pm 2\}}$, two to either side of $I_\mathrm{ref}$ in terms of image acquisition time. To this end, we assume that the input images are provided together with corresponding camera projection matrices $\mathrm{\mathbf{P}_\mathrm{i}} = \mathrm{\mathbf{K}} \left[\mathrm{\mathbf{R}^T_\mathrm{i}}\ -\mathrm{\mathbf{R}^T_\mathrm{i}}\mathrm{C_\mathrm{i}}\right]$, which consist of the rotation matrix $\mathrm{\mathbf{R}^T_\mathrm{i}} \in SO(3)$ and the position of the camera centers $\mathrm{C_\mathrm{i}} \in \mathbb{R}^3$, relative to a reference coordinate system. As we typically consider an image sequence, we assume the same intrinsic calibration matrix $\mathrm{\mathbf{K}}$ for all images.
%
%

The algorithm samples the scene by using multiple planes $\Pi_\mathrm{r} \in \mathbb{R}^3$, parameterized by their normal vector $\mathrm{n_\mathrm{r}}$ and distance $d_\mathrm{r}$ relative to $\mathrm{C_\mathrm{ref}}$. For each plane, the four matching images $I_\mathrm{ref\{\pm 1, \pm 2\}}$ are warped into $I_\mathrm{ref}$ according to the plane-induced homography 
\begin{equation}
\label{eq:homography}
\mathrm{\mathbf{H}^r_{i\rightarrow ref}} = \mathrm{\mathbf{K}} \cdot \frac{\mathrm{\mathbf{R}}-\mathrm{t}\cdot \mathrm{n_r}^\mathrm{T}}{d_r} \cdot \mathrm{\mathbf{K}}^{-1} , 
\end{equation}
\begin{tabbing} 
with \hspace{0.3cm} \= $\mathrm{\mathbf{R}}, \mathrm{t}$: \hspace{0.1cm} \= relative rotation and translation\\
\> \> between the cameras,\\
\> $\mathrm{n_\mathrm{r}}$: \> plane normal vector,\\
\> $d_\mathrm{r}$: \> plane distance from the reference camera.
\end{tabbing}\ \\
For each set of planes that share the same normal vector, we select the distances $d_r$ so that the scene is sampled inversely between two bounding planes $\Pi_\mathrm{min}$ and $\Pi_\mathrm{max}$ \citep{Ruf2017cross}. As similarity measure in the process of image matching, we choose a $9\times 7$ Census Transform \citep{Zabih1994}. In order to account for occlusions, we accumulate the matching cost within the left and right subset of the matching images and select the minimum of the two as suggested by \citet{Kang2001}. This yields a per-pixel matching cost for each plane which is stored in a three-dimensional cost volume $\mathcal{C}$ of size $W\times H\times D$, where $W$ and $H$ represent the image size and $D$ is the number of planes with which the scene is sampled.

Given the cost volume, we employ an edge-aware \gls*{SGM} optimization \citep{Hirschmueller2008} to extract the per-pixel minimum and with it for each pixel the distance $d_\mathrm{r}$ of the plane $\Pi_\mathrm{r}$ which best approximates the structure of the scene. The adapted energy function of the \gls*{SGM} optimization is as follows:
\begin{equation}
\label{eq:sgm}
\begin{aligned}
E(d_\mathrm{r}) = & \sum_\mathrm{p}{\mathcal{C}\left(\mathrm{p}, d_\mathrm{r}(\mathrm{p})\right)} + \\ & \sum_{\mathrm{q}\in N_\mathrm{p}}{P_1 \cdot \mathrm{T}\left[\left|\mathrm{Idx}( d_\mathrm{r}(\mathrm{p}))-\mathrm{Idx}\left( d_\mathrm{r}(\mathrm{q})\right)\right| = 1\right]} +  \\ & \sum_{\mathrm{q}\in N_\mathrm{p}}{P_2 \cdot \mathrm{T}\left[\left|\mathrm{Idx}\left( d_\mathrm{r}(\mathrm{p})\right)-\mathrm{Idx}\left( d_\mathrm{r}(\mathrm{q})\right)\right| > 1\right]} .
\end{aligned}
\end{equation}
With the first term, the matching costs of all pixels are accumulated for a given $d_\mathrm{r}$. With the second term, all neighborhood pixels $\mathrm{q}$ with a neighboring plane parameterization, i.e. the index of  $d_\mathrm{r}(\mathrm{p})$ and $d_\mathrm{r}(\mathrm{q})$ only changes by a maximum of one, are penalized with $P_1$. Neighboring pixels with other plane parameterizations are penalized with $P_2$. The optimization of Equation \ref{eq:sgm} is done by dynamic programming along eight concentric paths as suggested by \citet{Hirschmueller2008}.

In order to enforce discontinuities at object boundaries, we use the \gls*{LSD} introduced by \citet{Gioi2012} to create a binary line image $I^{line}_\mathrm{ref}$ of the reference image, where $I^{line}_\mathrm{ref}\left(\mathrm{p}\right) = 1$ indicates the existence of a line segment at a given pixel $\mathrm{p}$. With this line image, we adjust $P_2$ according to:
\begin{equation}
\label{eq:p2}
\begin{aligned}
P_2 &= 
	\begin{cases}
    	P_1 &,\ \text{if}\ I^{line}_\mathrm{ref}(\mathrm{p}) = 1 \\
        P_2 &,\ \text{otherwise}.
    \end{cases}
\end{aligned}
\end{equation}
Thus we reduce $P_2$ to $P_1$, if a given pixel $\mathrm{p}$ is on a line segment of $I^{line}_\mathrm{ref}$, allowing neighboring pixels with strong discontinuities to influence the selection of the optimal $d_\mathrm{r}$. Dependent on the parameterization of the \gls*{LSD}, most line segments are detected at object boundaries, which allows us to enforce strong edges in the depth image.

Having found the optimal per-pixel plane parameterization $\hat{\Pi}\left(\mathrm{p}\right)$, the final depth image is calculated by intersecting a viewing ray through pixel $\mathrm{p}$ with $\hat{\Pi}\left(\mathrm{p}\right)$. 

For a selective reconstruction that is confined to identified \gls*{ROIs}, we perform the per-plane image warping on the complete images, yet the matching and the subsequent \gls*{SGM} optimization is only done within the given \gls*{ROIs}, treating each bounding box individually. We optimized our algorithm with CUDA, achieving real-time performance when aiming to estimate depth maps for every key-frame which are typically generated at 1\,Hz - 2\,Hz by state-of-the-art SLAM systems. This enables the online analysis of the depicted scene.

\section{EXPERIMENTS}
\label{sec:eval}

\sloppy

Before presenting the results achieved by the Faster R-CNN and our algorithm for depth estimation, we introduce the datasets used for training and evaluation. 

\subsection{Datasets}
\label{sec:eval_datasets}

\begin{figure*}[!ht]
  \centering
  \subfigure[]{\includegraphics[width=0.68\columnwidth]{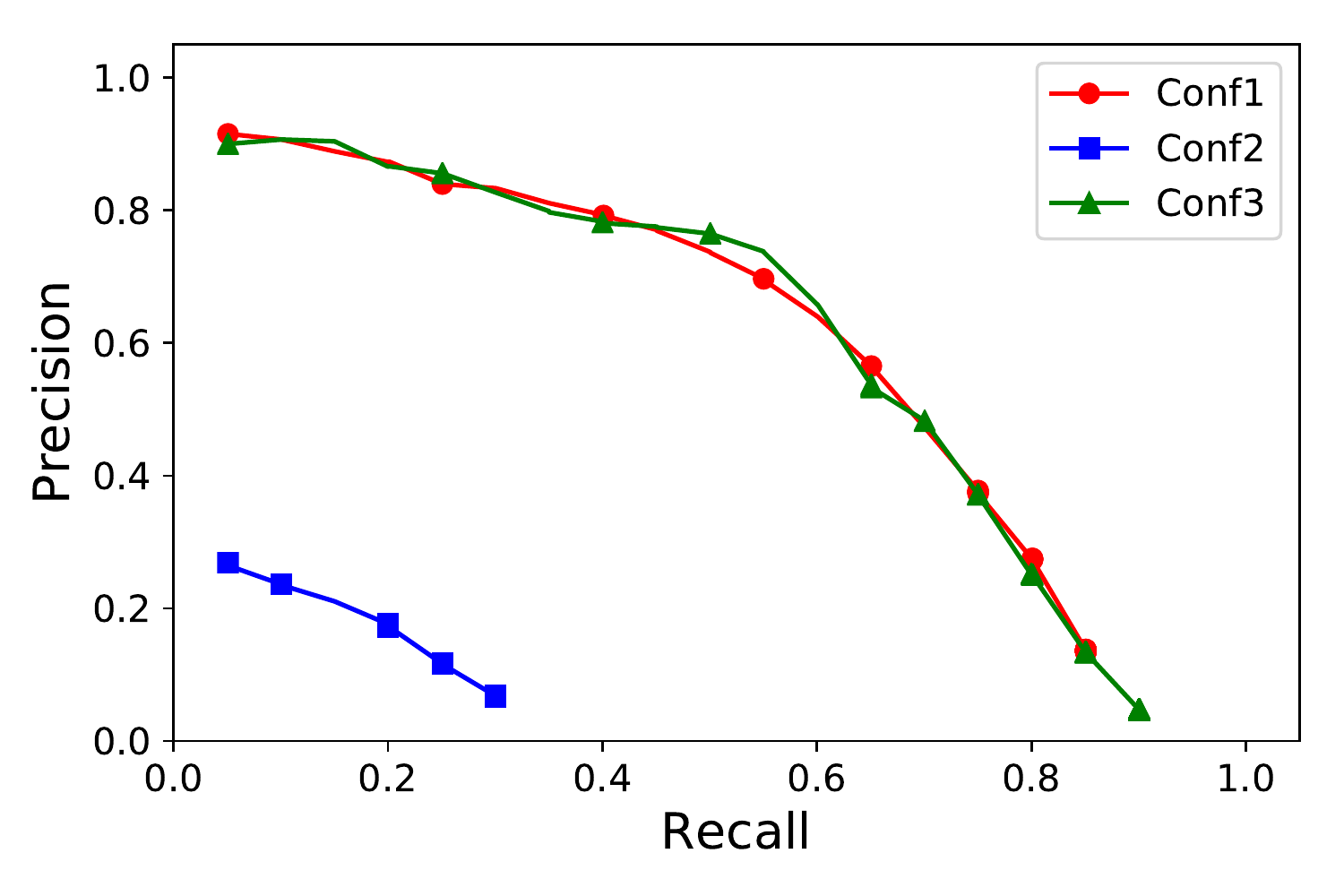}\label{fig:precRecall_valCS}} \hfill
  \subfigure[]{\includegraphics[width=0.68\columnwidth]{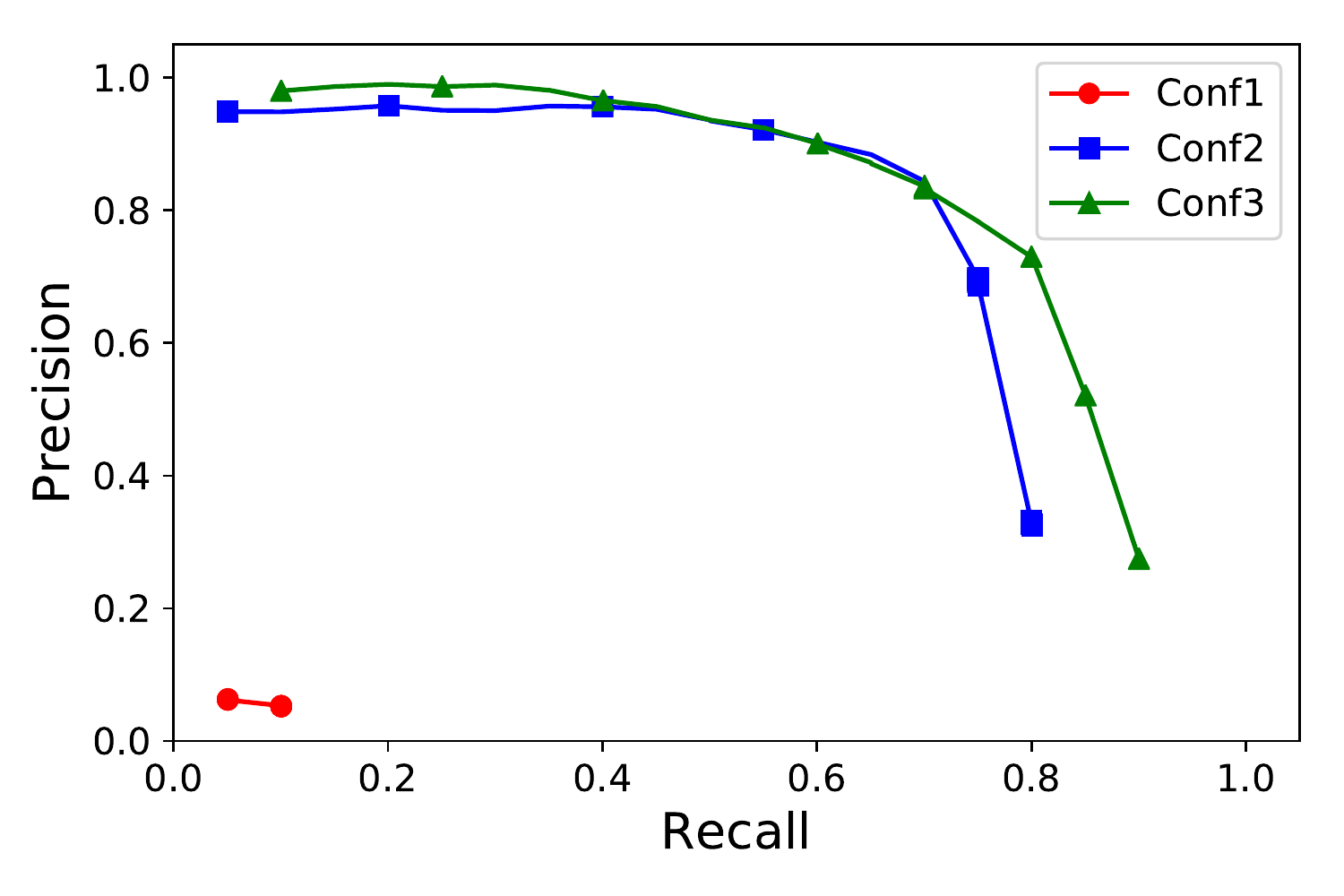}\label{fig:precRecall_valISPRS}} \hfill
  \subfigure[]{\includegraphics[width=0.68\columnwidth]{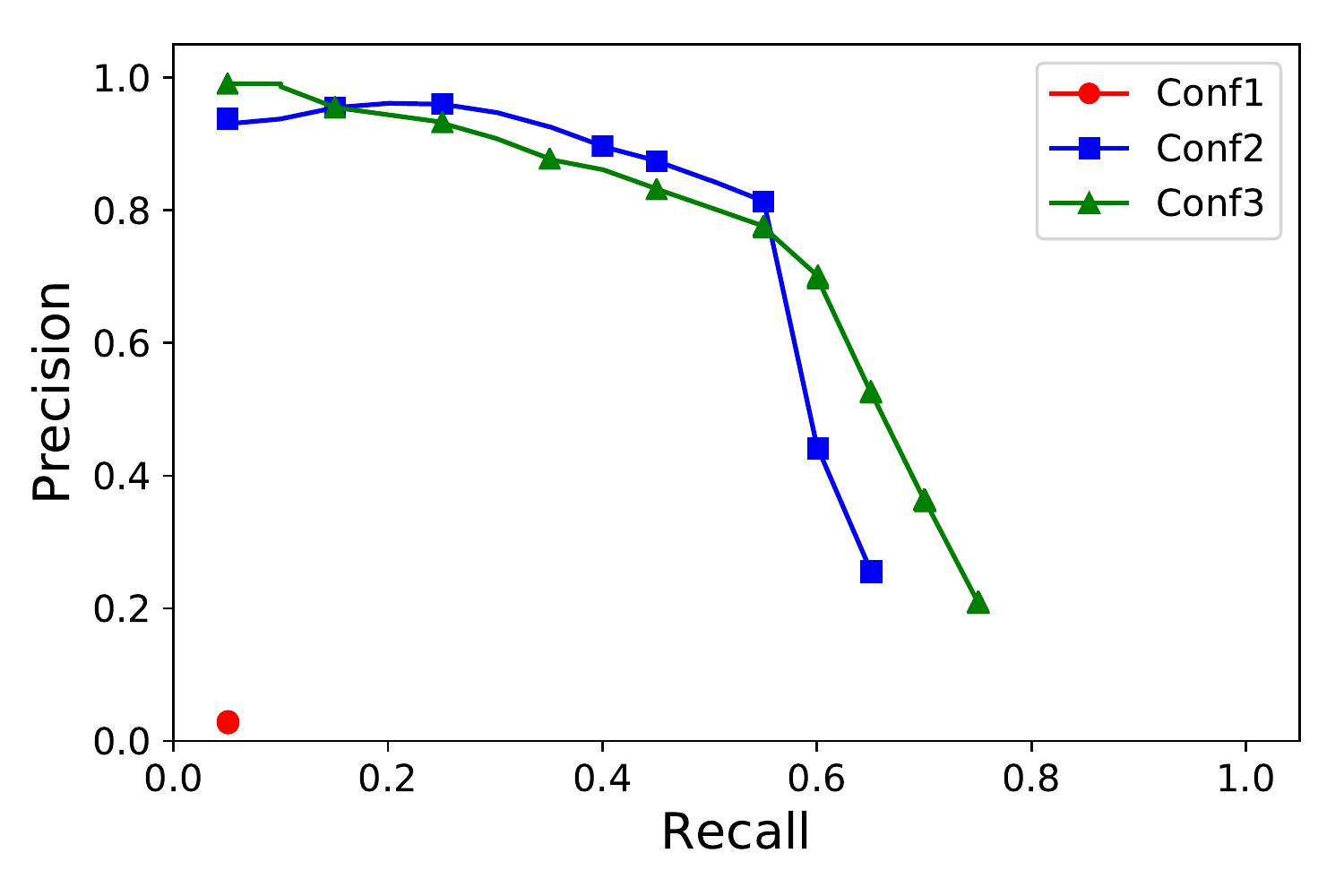}\label{fig:precRecall_valGoogle}} \hfill
  \caption{Precision-recall curves for different validation datasets. \textbf{(a)} Results achieved by the three training configurations evaluated on the Cityscapes \textit{validation} set. \textbf{(b)} Results achieved on the ISPRS \textit{validation} set. \textbf{(c)} Precision-recall values corresponding to an evaluation on the GoogleEarth dataset.}
  \label{fig:precision_recall}
\end{figure*}

Deep learning based approaches, such as the Faster R-CNN, require a large amount of training data. To the best of our knowledge, there exists no suitable training dataset for object detection from oblique aerial imagery. Assuming that such images depict both the fa\c{c}ade and the roof of buildings, we have adopted two different datasets for training the \gls*{CNN}, namely the Cityscapes dataset for semantic urban scene understanding \citep{Cordts2016Cityscapes} and the dataset given with the ISPRS benchmark for multi-platform photogrammetry \citep{Nex2015Isprs}. 

The Cityscapes dataset (cf. Figure \ref{fig:datasets_cityscape}) consists of a large number of images which were captured from a car driving through urban areas. It provides a pixel-wise semantic ground truth. 
We generated a ground truth for building extraction by converting the semantic labeling of the building class into axis-aligned bounding boxes, needed for the training of the Faster R-CNN. The dataset provides disjoint subsets used for training and evaluation. 
The \textit{training} subset consists of approx.\ 2900 images, whereas the \textit{validation} subset contains 500 images.

The Dortmund-Zeche-Zollern subset of the ISPRS benchmark (cf. Figure \ref{fig:datasets_isprs}) consists of few large images with approx.\ $48$ mega pixels captured from nadir and oblique viewpoints. 
Due to the large image size, we have cropped each input image into 64 equal-sized subimages with a size of $1022\times 766$ pixels. Because this benchmark does not contain any annotated ground truth, we have selected 700 oblique images and annotated the depicted buildings with bounding boxes. We have divided the annotated images into a \textit{training} and a \textit{validation} subset containing 500 and 200 images, respectively. 

To further evaluate the performance of Faster R-CNN to detect buildings on a different dataset, we have generated a semi-synthetic dataset from GoogleEarth (cf. Figure \ref{fig:datasets_google}). We have selected seven different scenes with underlying model data in order to render different viewpoints. The scenes differ in the amount and the kind of buildings they show. We have rendered each scene with three off-nadir angles ($0^{\circ}$, $30^{\circ}$ and $60^{\circ}$) and from five distances (100\m, 150\m, 200\m, 250\m and 300\m), where each image has a size of $1920\times 1080$ pixels. 

A fourth dataset was used to demonstrate the performance of our approach on a real-world scenario (cf. Figure \ref{fig:datasets_tmb}). It represents an own dataset of a single building which was captured with a DJI Phantom 3 from three different heights (2\m, 8\m and 15\m) while flying around the front of the building. This dataset fits to the use-case of a selective 3d reconstruction as the building only covers a small amount of the input images. The input images are of size $1920\times 1080$ pixels.

\subsection{Building Extraction from Aerial Imagery}
\label{sec:eval_detection}

\subsubsection{Training}
\label{sec:eval_training}

We have selected three configurations to train the Faster R-CNN for building extraction. The configurations differ in the training datasets used:
\begin{description}[topsep=0pt, partopsep=0pt, itemsep=0pt]
	\item \textit{Conf1}: Cityscapes
	\item \textit{Conf2}: ISPRS Dortmund-Zeche-Zollern
	\item \textit{Conf3}: Cityscapes + ISPRS Dortmund-Zeche-Zollern
\end{description}
For each dataset, the available \textit{training} subset was used. As a training method of the Faster R-CNN, we have chosen the presented \textit{approximate joint training} which alternates between training the Fast R-CNN network and the additional \gls*{RPN}. 
\citet{Ren2017FasterRCNN} suggest that it results in an approx.\ $1.5\times$ speedup and produce a similar accuracy as when using the \textit{alternating training} method. For all configurations, we have initialized the weights with a VGG-16 model which was pretrained on the ImageNet dataset. We have used 250{,}000 training iterations.

In order to increase the amount of training data and to oppose overfitting, we have augmented the initial training sets with affinely transformed copies of the input images. First, we horizontally flipped the input data to increase the database and generalize the model. Second, due to the perspective of the Cityscapes image data, most bounding boxes surrounding buildings are of slim horizontal nature. In order to oppose overfitting to such geometry, we have rotated the images by 90$^{\circ}$ and added them to the dataset. Third, to accommodate for different object sizes, the input images were randomly downsampled by three different scaling factors. As the image size of the Cityscapes and the ISPRS datasets differ, we used different scaling factors for each dataset. In terms of the Cityscapes dataset, we have downsampled the image data to 60\%, 40\% or 20\% of its original size. For the ISPRS dataset, we use images scaled to 100\%, 80\% or 60\% of their original image size.

\subsubsection{Results}
\label{sec:eval_building_results}

\begin{table}[!t] 
\caption{Quantitative results achieved by the three configurations evaluated on the three validation datasets. The underlined values represent the baselines as the training and evaluation was done on the same kind of data.}
\label{tab:quantEvalSoftNMS}   
\centering
\begin{tabular}{|l|l|l|l|}\hline
       			& \textit{Conf1} & \textit{Conf2} & \textit{Conf3} \\ 
       \hline \hline
       \multirow{ 2}{*}{\diagbox{Eval.}{Train.}}	& Cityscapes & ISPRS		& Cityscapes 	\\ 
      		    									& 			& 	 		& + ISPRS			\\ 
       \hline
       Cityscapes 									& \underline{60.57} \%	& 6.62 \% 	& \textbf{60.76} \%		\\ 
  		\hline
       ISPRS										& 1.01 \% 	& \underline{73.43} \%	& \textbf{80.81} \%		\\ 
       	\hline
       GoogleEarth 										& 0.45 \%	& 56.81 \% 	& \textbf{60.70} \% 		\\ 
       \hline
\end{tabular}
\end{table}

\begin{figure*}[!ht]
  \centering
  \subfigure[]{\includegraphics[width=0.67\columnwidth]{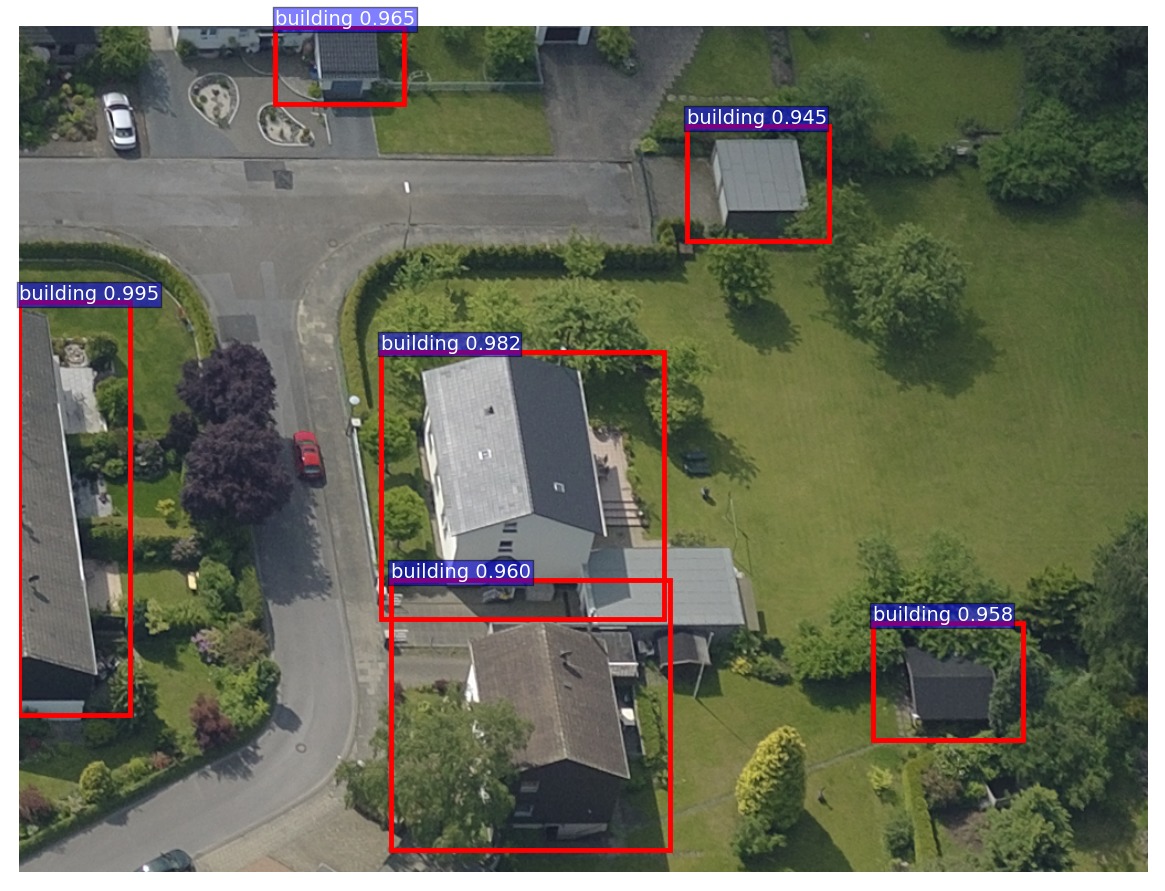}\label{fig:qual_detection_a}} \hfill
  \subfigure[]{\includegraphics[width=0.67\columnwidth]{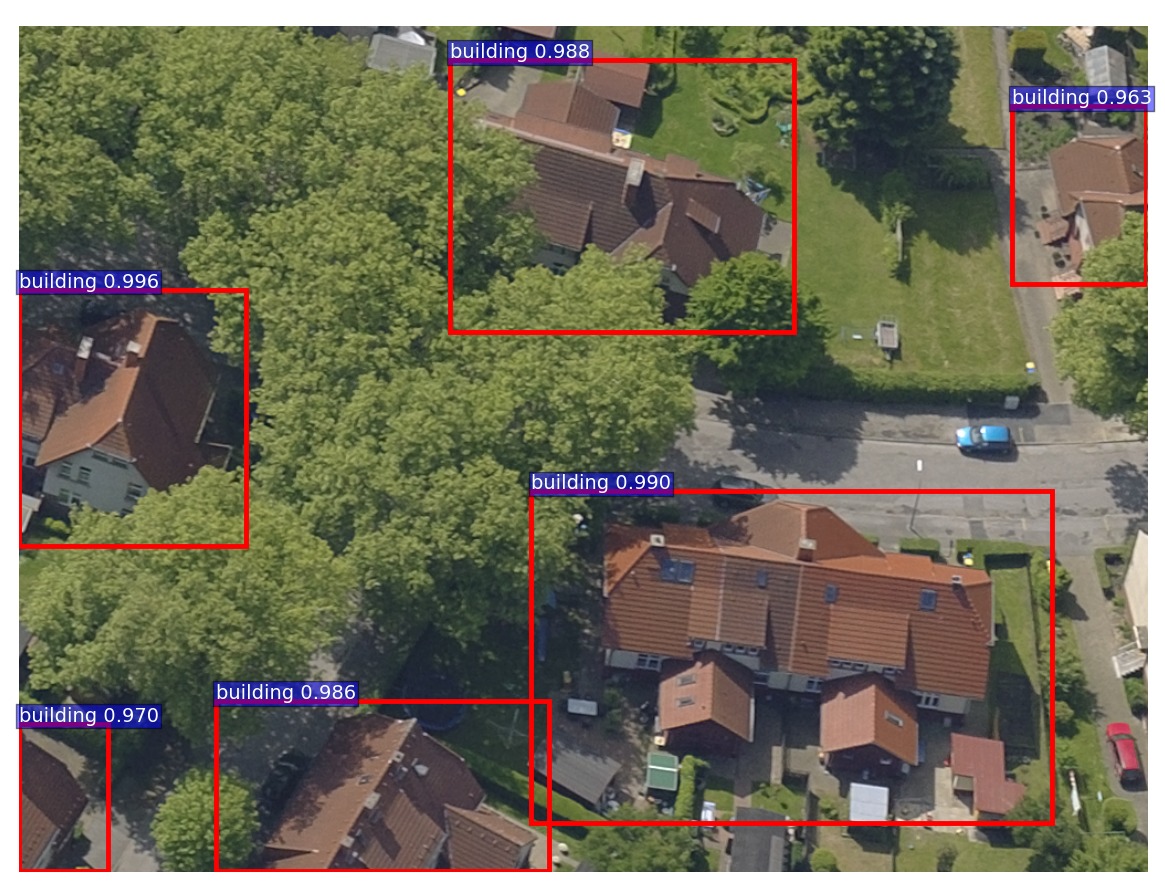}\label{fig:qual_detection_b}} \hfill
  \subfigure[]{\includegraphics[width=0.67\columnwidth]{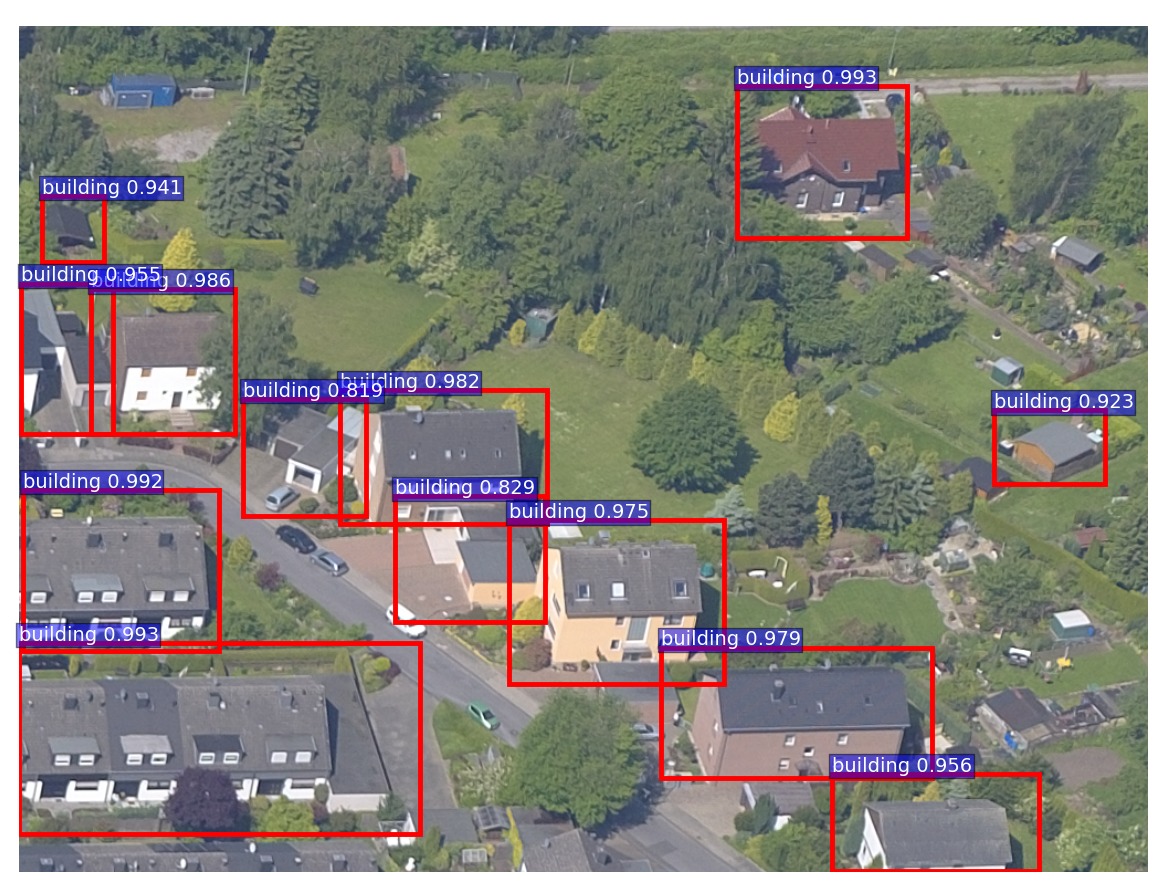}\label{fig:qual_detection_c}} \\  \vspace{-2mm}
  \subfigure[]{\includegraphics[width=0.67\columnwidth]{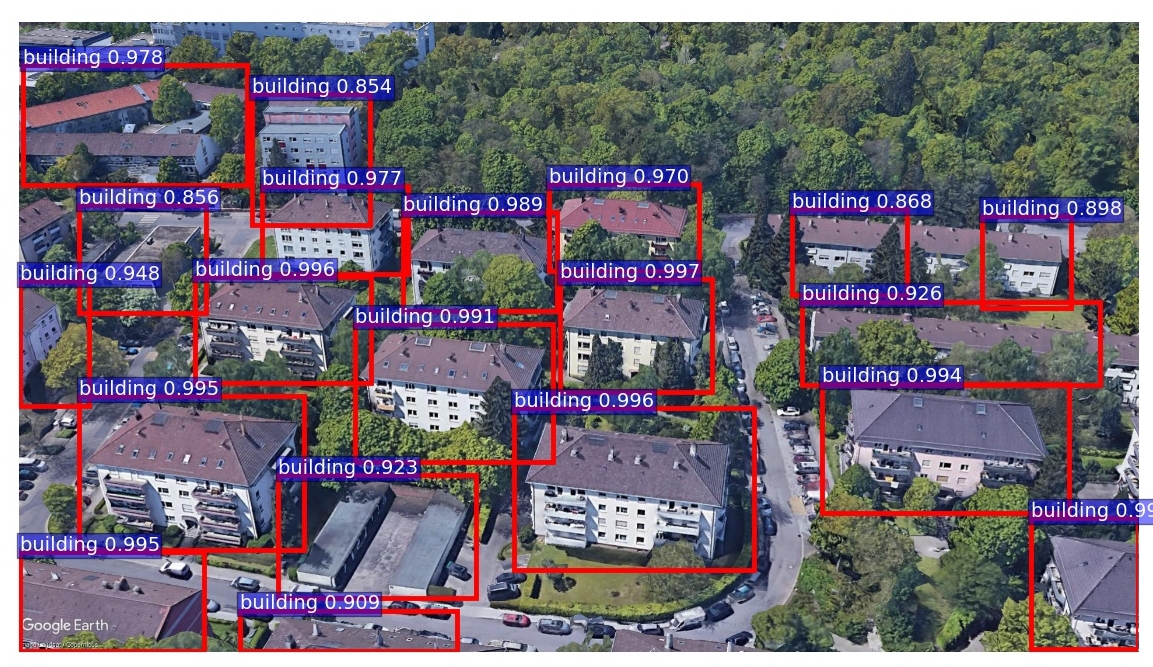}\label{fig:qual_detection_d}} \hfill
  \subfigure[]{\includegraphics[width=0.67\columnwidth]{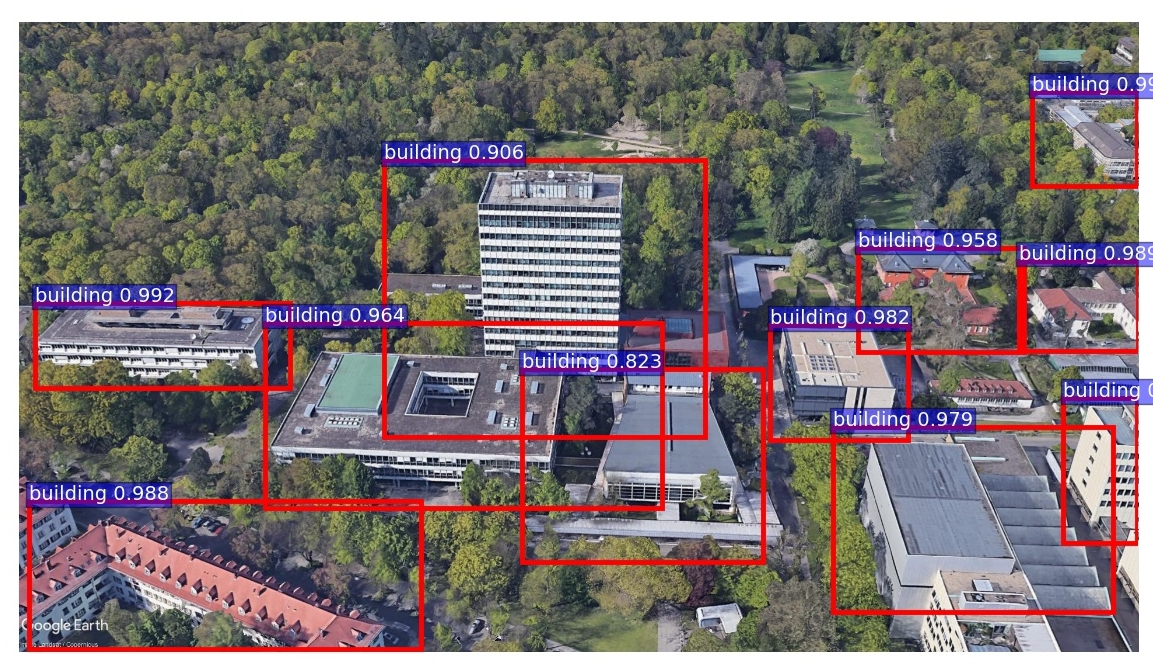}\label{fig:qual_detection_e}} \hfill
  \subfigure[]{\includegraphics[width=0.67\columnwidth]{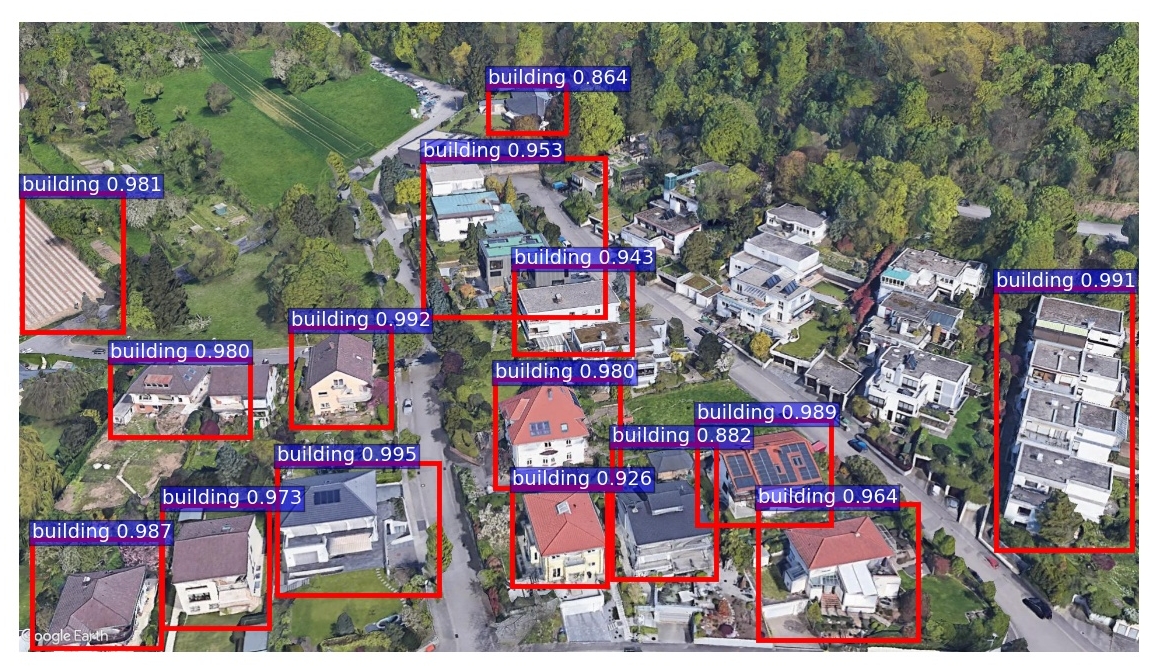}\label{fig:qual_detection_f}} 
  \caption{Excerpt of the qualitative results achieved by the Faster R-CNN trained with \textit{Conf3}. Figures (a), (b) and (c) depict results from the ISPRS dataset. Figures (d), (e) and (f) show an excerpt of the detection from the GoogleEarth dataset. Only detections with a confidence score $>0.8$ are highlighted.}
  \label{fig:qual_detection}
\end{figure*}
\begin{figure*}[!ht]
  \centering
  \subfigure[]{\includegraphics[width=0.67\columnwidth,frame]{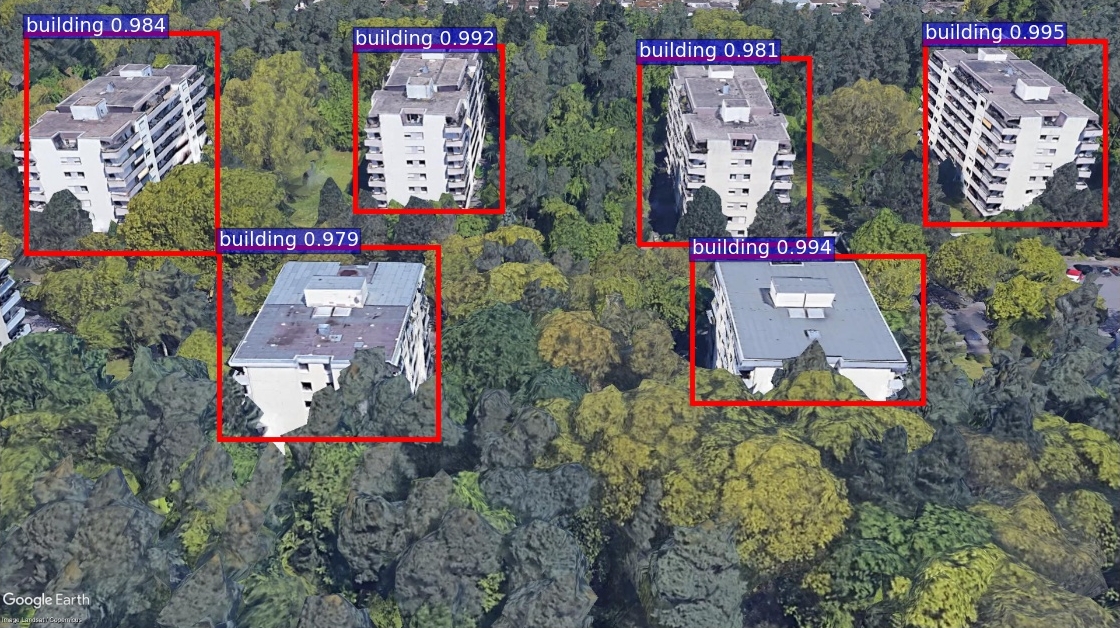}\label{fig:qual_depth_a}} \hfill
  \subfigure[]{\includegraphics[width=0.67\columnwidth,frame]{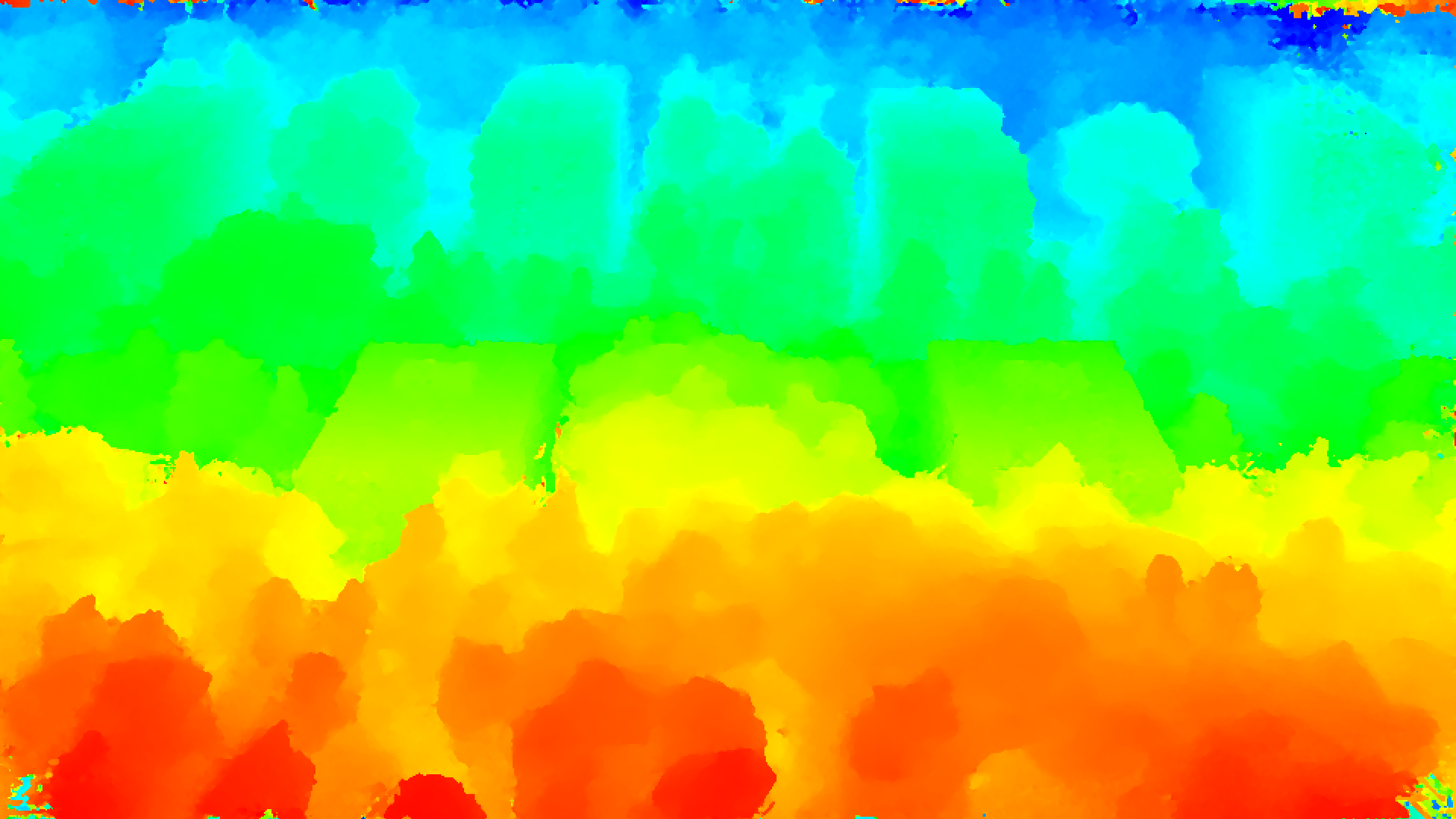}\label{fig:qual_depth_b}} \hfill
  \subfigure[]{\includegraphics[width=0.67\columnwidth,frame]{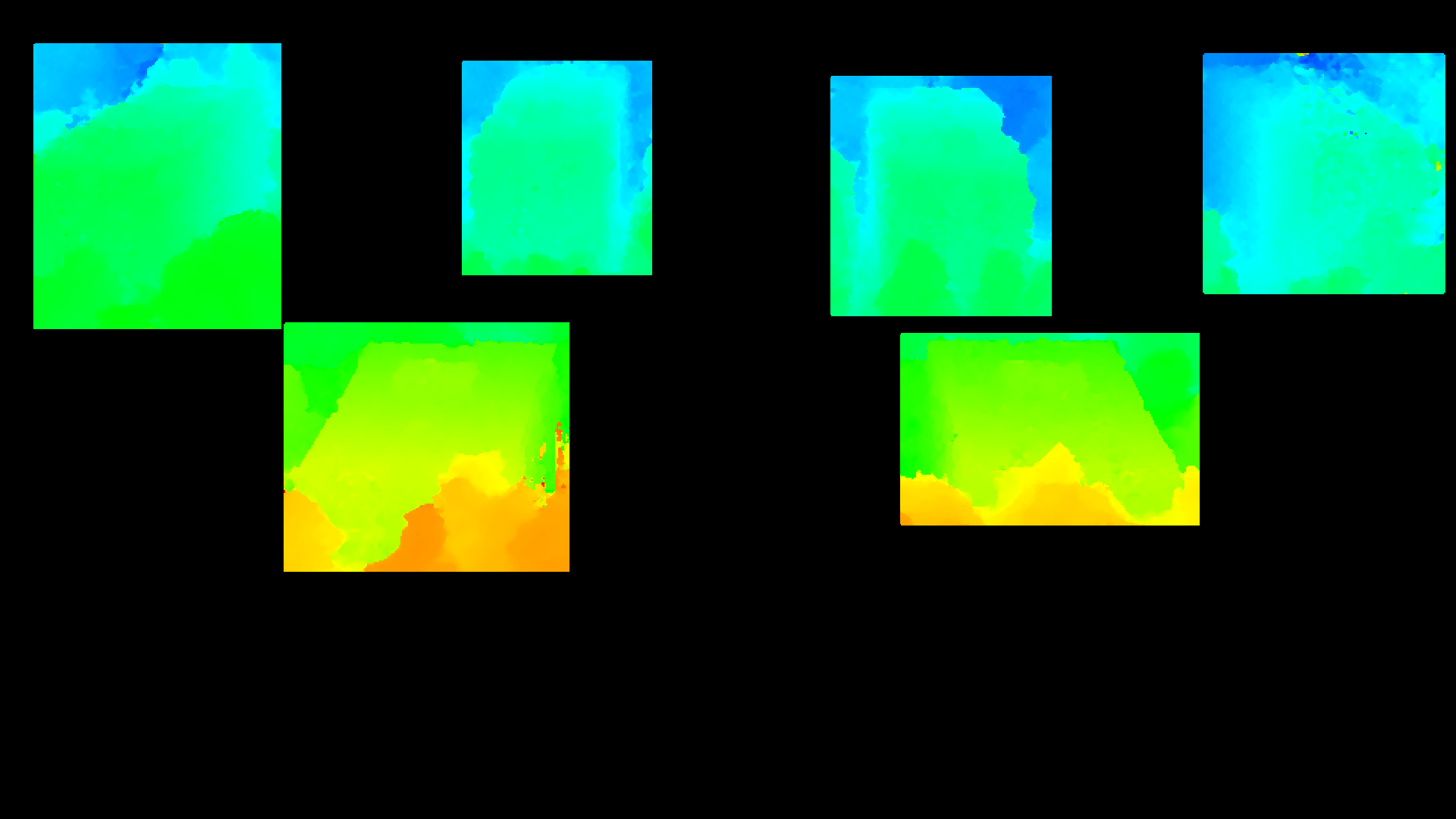}\label{fig:qual_depth_c}} \\ \vspace{-2mm}
  \subfigure[]{\includegraphics[width=0.67\columnwidth,frame]{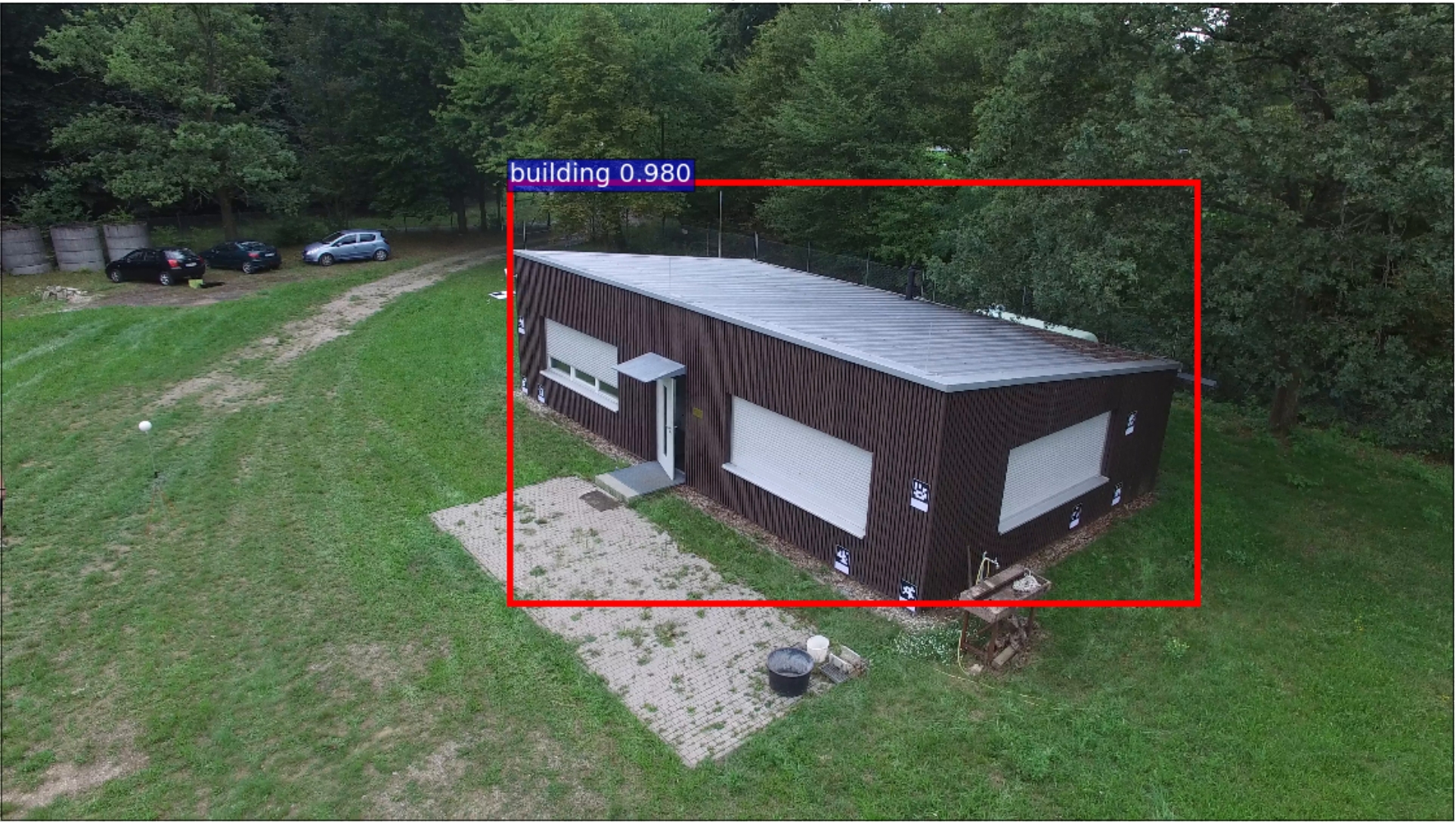}\label{fig:qual_depth_d}} \hfill
  \subfigure[]{\includegraphics[width=0.67\columnwidth,frame]{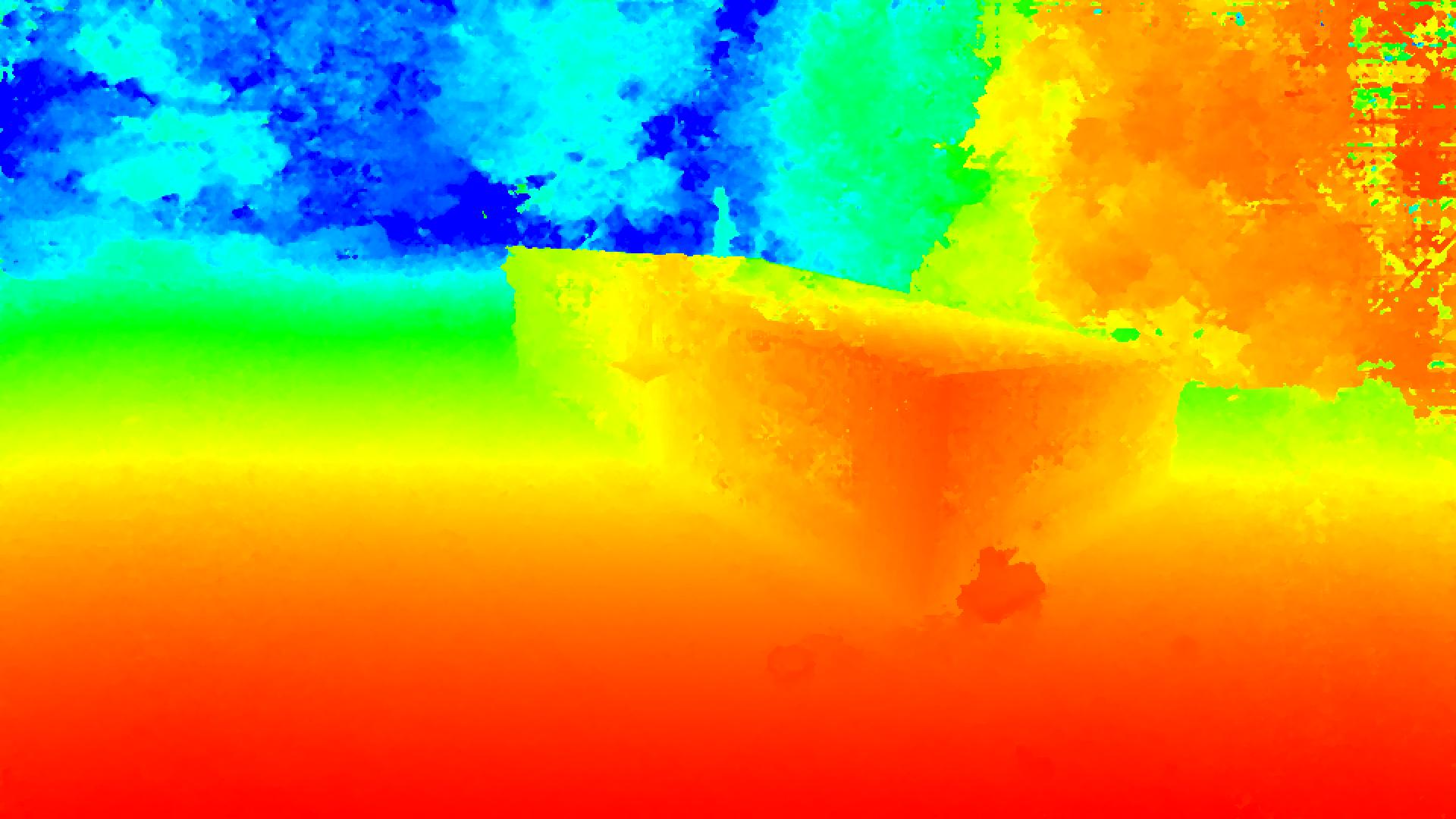}\label{fig:qual_depth_e}} \hfill
  \subfigure[]{\includegraphics[width=0.67\columnwidth,frame]{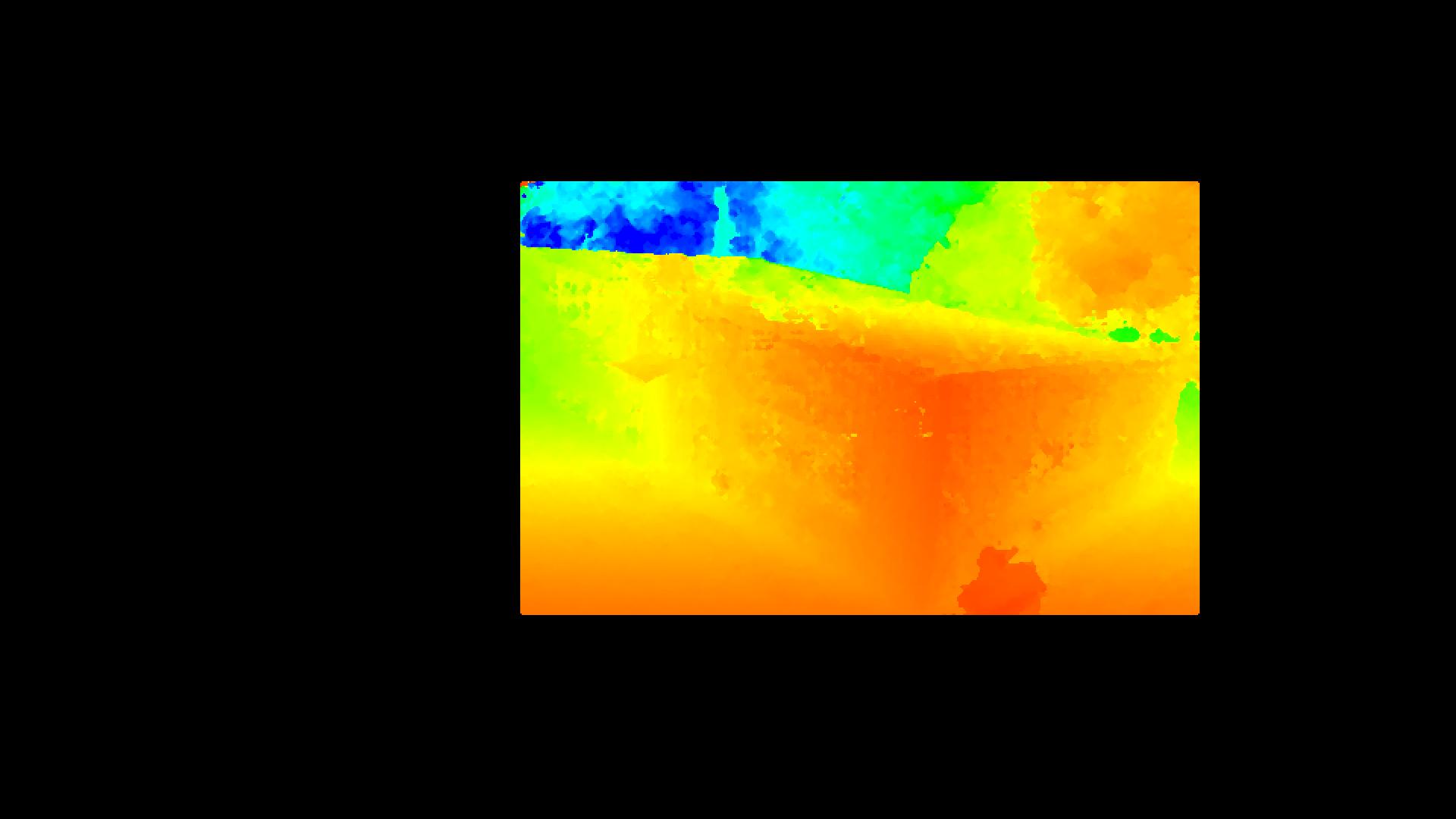}\label{fig:qual_depth_f}}
  \caption{Results of selective image-based depth estimation. Figures (a) and (d) show the reference frames with corresponding detections. Figure (a) was captured at a distance of 150\m with an off-nadir angle of 60$^{\circ}$. Figure (d) was captured at an altitude of 15\m. Full depth maps corresponding to the reference frames are shown in Figures (b) and (e). The results of a selective image-based depth estimation are shown in Figures (c) and (f), where the depth is only computed for the detected objects. Depth values are color-coded going from red (near) to blue (far).}
  \label{fig:qual_depth}
\end{figure*}

We have evaluated the performance of the Faster R-CNN to detect buildings on the \textit{validation} subsets of the Cityscapes and ISPRS datasets as well as the generated GoogleEarth dataset. We have evaluated models of the Faster R-CNN that were trained with the three previously described training configurations. For evaluation on the Cityscapes dataset, we have downscaled the test images to 40\% of their original size. The test images of the ISPRS dataset were used with the original image size, while the GoogleEarth images were scaled to 80\%. These values were determined empirically, giving the best results for each dataset. We have set the \gls*{RPN} to generate 12{,}000 proposals for each image.

Table \ref{tab:quantEvalSoftNMS} shows the \gls*{AP} achieved by the three training configurations in detecting buildings on the three evaluation datasets. 
For the Soft-\gls*{NMS}, we have used a linear decay function with a \gls*{NMS} threshold of 0.3.
Furthermore, in Figure \ref{fig:precRecall_valCS} we have plotted the precision-recall-curves of \textit{Conf1-3}, evaluated on the \textit{validation} set of the Cityscapes dataset. 
Figure \ref{fig:precRecall_valISPRS} shows the performance of the training configurations \wrt the ISPRS \textit{validation} dataset and Figure \ref{fig:precRecall_valGoogle} shows the results of the evaluation on the GoogleEarth dataset.

Figure \ref{fig:qual_detection} shows an excerpt of the results achieved by the Faster R-CNN trained with \textit{Conf3} and combined with Soft-\gls*{NMS}. This configuration was chosen as it quantitatively achieves the best results on all three datasets. While Figures \ref{fig:qual_detection_a}~-~\ref{fig:qual_detection_c} hold the results for the ISPRS dataset, Figures \ref{fig:qual_detection_d}~-~\ref{fig:qual_detection_f} show the detections for images of the GoogleEarth dataset. Each detection is marked with a red bounding box. With it, the Faster R-CNN computes a confidence score $\left[0.0,1.0\right]$ with which it identifies an object as a building. These scores are displayed above each \gls*{ROI} together with the corresponding class name of the detected object. In Figure \ref{fig:qual_detection}, only objects with a score $> 0.8$ are displayed.

All training and experiments were performed on a NVIDIA Titan X. The runtime to process one image for the Cityscapes dataset is approx.\ 140\,ms, for the ISPRS dataset approx.\ 160\,ms and for the GoogleEarth dataset approx.\ 285\,ms. 
The differences arise from the different image sizes which were used for the experiment.

\subsection{Selective Image-based Depth Estimation}
\label{sec:eval_depth}

We have evaluated our approach for selective image-based depth estimation on two datasets. First, we have generated an image set, made up of five different viewpoints around a reference image (cf. Figure \ref{fig:qual_depth_a}), from one of the scenes of the GoogleEarth dataset. The scene shows a few isolated buildings surrounded by lots of vegetation, which is one of the key motivations to use an object detection algorithm to extract \gls*{ROIs} prior to the depth estimation. 

Further experiments are done on imagery taken from our own dataset (cf. Figures \ref{fig:datasets_tmb} and \ref{fig:qual_depth_d}). 
Again, it shows an isolated object of interest surrounded by lots of scenery not relevant for the task of building reconstruction. 
It was captured from 15\m altitude.

We have parameterized our algorithm to use 128 frontoparallel planes. The penalties of our \gls*{SGM} optimization were set to $P_1~=~5$ and $P_2~=~50$ in order to enforce smooth reconstructions within object boundaries. The \gls*{LSD} was used with the presented default configurations and the input images were downsampled to half of their size. The camera projection matrices were computed using multicore bundle adjustment \citep{Wu2011multicore}. We employ a final $3\times 3$ median filter to reduce outliers.

Figure \ref{fig:qual_depth} holds an excerpt of our experiments for selective depth estimation. Figures \ref{fig:qual_depth_a} and \ref{fig:qual_depth_d} show the respective reference images with the \gls*{ROIs} generated by the Faster R-CNN. Again, we use the model trained with \textit{Conf3}. In Figures \ref{fig:qual_depth_b} and \ref{fig:qual_depth_e}, the full depth maps are shown, in which the depth values are visually encoded, going from red (near) to blue (far). In Figures \ref{fig:qual_depth_c} and \ref{fig:qual_depth_f}, the results of the selective depth estimation are shown. Hereby, the depth values were only computed for the \gls*{ROIs} identified by the Faster R-CNN and fused together into one depth map corresponding to $I_\mathrm{ref}$.

Again, all experiments were performed on a NVIDIA Titan~X. The depth estimation for the full image 
took approx.\ 510\,ms. The selective depth estimation reduced the complete runtime to an average of 226\;ms.

\section{DISCUSSION}
\label{sec:discussion}

\sloppy

We consider the underlined numbers in Table \ref{tab:quantEvalSoftNMS} as baseline, because they represent the performance of the Faster R-CNN in detecting buildings when respectively trained and evaluated on the same kind of data. 
Concerning these baselines, we achieve equal or better results when training the Faster R-CNN with a combination of the Cityscapes and the ISPRS datasets, i.e. \textit{Conf3}. 
Even when evaluated on a dataset, which it has not seen during training, i.e. the GoogleEarth dataset, the \textit{Conf3} model achieves equal results as the \textit{Conf1} model when evaluated on the Cityscapes dataset. 

The loss in performance \wrt the GoogleEarth dataset can be attributed to different factors. First, the fact that it is a different dataset which was not used during training. We assume that the model does not generalize well to building structures which are atypical \wrt the training data. Figure \ref{fig:qual_detection_f} shows that the model does not identify the building blocks on the right, as well as the individual buildings in the center. Furthermore, the field on the left is falsely detected as a building with a significantly high score. A second factor may be the quality of the underlying 3d model of GoogleEarth and the projected textures. A poor quality in the model or the texture may result in false detections. Nonetheless, the results achieved by \textit{Conf3} are significantly better than the ones achieved by the other models on this dataset.

The precision-recall curves reveal how the different training datasets influence the results achieved. Figure \ref{fig:precRecall_valCS} shows that due to the size of the Cityscapes dataset, in comparison to the annotated subset of the ISPRS dataset, the \textit{Conf3} model delivers very similar results as the \textit{Conf1} model. 
Figure \ref{fig:precRecall_valISPRS} shows that the additional use of the Cityscapes dataset improves the results compared to the baseline of the ISPRS dataset, i.e. \textit{Conf2}. Nonetheless, in order to detect buildings from aerial imagery, the use of the ISPRS training data is needed, as Figures \ref{fig:precRecall_valISPRS} and \ref{fig:precRecall_valGoogle} show that \textit{Conf1} does not achieve any reasonable results.

The training process of the Faster R-CNN keeps the first two convolutional layers of the VGG-16 network fixed \citep{Girshick2015FastRCNN}. Changing this, so that the weights of all layers are trained worsens the results by 1\%-2\%. A similar effect was observed when a dense-sparse-dense training \citep{Han2017dsd} was performed in order to regularize the model. In further experiments, we have evaluated the use of standard \gls*{NMS} instead of Soft-\gls*{NMS}. 
As expected, the results achieved were inferior to those presented.

The results in Figure \ref{fig:qual_depth} show that our algorithm for image-based depth estimation allows to reconstruct building structures while preserving sharp edges at object boundaries. Our algorithm achieves real-time performance for the full reconstruction. Nonetheless, the runtime can be reduced by 50\% if only objects of interest are considered. The results also reveal that there seams to be no significant loss in quality of the selective reconstruction compared to the full depth map. As our main focus in this paper is the use of Faster R-CNN to extract buildings from aerial imagery, an elaborate performance evaluation of our image-based depth estimation algorithm is done in other work.

As the building detection currently only extracts axis-aligned \gls*{ROIs}, it happens that parts of irrelevant objects are encapsulated in the bounding boxes. To overcome this, one can use a semantic segmentation that attributes a specific object class to each pixel. However, as our depth estimation algorithm is parallelized on the GPU, it needs to be invoked on a regular grid. The recently published Mask R-CNN \citep{He2017MaskRCNN} would allow an extraction of bounding boxes prior to reconstruction and a pixel-wise post-filtering of the resulting depth images.

Altogether, this paper shows that we can use transfer learning of the Faster R-CNN for building extraction from oblique aerial imagery, by using a combination of a large ground-based dataset and a smaller aerial dataset. The results show that the use of the large Cityscapes dataset is necessary to learn building-related features. Nonetheless, a small aerial training dataset is vital to generalize the model to perform detections in aerial imagery. With our approach, we achieve a real-time extraction and selective reconstruction of building structures. However, a registration of the reference frame \wrt the city model is needed in order to perform a 3d change detection and analysis.

\section{CONCLUSION \& FUTURE WORK}
\label{sec:conclusion}

\sloppy

In summary, we present a methodology for cross-domain building extraction from aerial imagery with \gls*{CNNs}. We combine a large ground-based dataset for urban scene understanding with a smaller number of images from an aerial dataset to train the Faster R-CNN method for real-time deep object detection. We achieve an \gls*{AP} of about 80\% for the task of building extraction on a selected evaluation dataset. In combination with a presented algorithm for real-time image-based depth estimation, it allows us to selectivly reconstruct buildings from aerial imagery, preserving sharp edges. These reconstructions can be used for 3d scene analysis or change detection in city models.

As axis-aligned bounding boxes do not always appropriately approximate the building outlines, we aim to incorporate a semantic segmentation for post-filtering the depth map. To this end, in the image-based depth estimation, we perform the same image warpings for all identified \gls*{ROIs}. In future work, different homographies are to be used for each \gls*{ROI}, which allows to further adjust the sampling process to the object of interest. Finally, we plan to combine it with an algorithm to register the image data against textureless model data, allowing us to match the depth map against the model data for structural change detection. 

\section*{ACKNOWLEDGEMENTS}
\label{sec:acknowledgments}

The authors would like to acknowledge the provision of the datasets for multi-platform photogrammetry by ISPRS and EuroSDR, released in conjunction with the ISPRS scientific initiative 2014 and 2015, led by ISPRS ICWG I/Vb.

{\footnotesize 
	\begin{spacing}{0.9}
    \setlength{\bibsep}{2pt}
		\bibliography{deep-cross-domain_biblio} 
	\end{spacing}
}







\end{document}